\documentclass[conference]{IEEEtran}
\IEEEoverridecommandlockouts

\usepackage{float}
\usepackage{cite}
\usepackage{soul,color}
\usepackage{amsmath,amssymb,amsfonts}
\usepackage{algorithm}
\usepackage{algpseudocode}
\usepackage{graphicx}
\usepackage{textcomp}
\usepackage{xcolor}
\usepackage[font=small,skip=0pt]{caption}
\algnewcommand{\LineComment}[1]{\State \(\triangleright\) #1}
\def\BibTeX{{\rm B\kern-.05em{\sc i\kern-.025em b}\kern-.08em
    T\kern-.1667em\lower.7ex\hbox{E}\kern-.125emX}}
    
\begin{document}

\title{Online Informative Sampling using Semantic Features in Underwater Environments
}

\author{\IEEEauthorblockN{Shrutika Vishal Thengane \qquad Yu Xiang Tan \qquad Marcel Bartholomeus Prasetyo \qquad Malika Meghjani}
}

\author{Shrutika Vishal Thengane, Yu Xiang Tan, Marcel Bartholomeus Prasetyo, \\ and Malika Meghjani
    \thanks{Shrutika Vishal Thengane, Yu Xiang Tan, Marcel Bartholomeus Prasetyo and Malika Meghjani are with Singapore University of Technology and Design, Singapore. {\tt\small yuxiang\_tan@mymail.sutd.edu.sg, \{shrutika\_thengane, marcel\_prasetyo, malika\_meghjani\}@sutd.edu.sg}} %
}

\maketitle

\begin{abstract}
The underwater world remains largely unexplored, with Autonomous Underwater Vehicles (AUVs) playing a crucial role in sub-sea explorations. However, continuous monitoring of underwater environments using AUVs can generate a significant amount of data. In addition, sending live data feed from an underwater environment requires dedicated on-board data storage options for AUVs which can hinder requirements of other higher priority tasks. Informative sampling techniques offer a solution by condensing observations. In this paper, we present a semantically-aware online informative sampling (ON-IS) approach which samples an AUV's visual experience in real-time. Specifically, we obtain visual features from a fine-tuned object detection model to align the sampling outcomes with the desired semantic information. Our contributions are (a) a novel Semantic Online Informative Sampling (SON-IS) algorithm, (b) a user study to validate the proposed approach and (c) a novel evaluation metric to score our proposed algorithm with respect to the suggested samples by human subjects.
\end{abstract}

\vspace{-0.01\textheight}
\section{Introduction}

Visual informative sampling algorithms help to identify distinctive temporal instances. This is useful for Autonomous Underwater Vehicles (AUVs) performing tasks such as deep sea exploration or coral reef monitoring as both require long-term observations with human-in-loop. For monitoring tasks, the informative sampling algorithms can significantly reduce the amount of content to be analyzed offline, thereby improving the task efficiency for the human operator. This is done by using informative sampling as a filter to identify the significant events in the video. For underwater exploration tasks, an online visual informative sampling algorithm can guide the robot to collect new and unexplored data.
In a scenario with low-bandwidth communication for remote operations between the robot and the surface, the online sampling could also be performed on the robot to succinctly transfer crucial information to the surface for quick understanding of the situation.

Informative sampling techniques can be categorized into two distinct groups: (a) Offline Informative Sampling (OFF-IS) and (b) Online Informative Sampling (ON-IS). In OFF-IS algorithms \cite{otani2016video}, the entire visual data and its duration are available to the algorithm in advance, and these algorithms compute features for all frames all at once. These features are then grouped based on similarities, and the sampling process involves selecting a representative frame from each group. In contrast, ON-IS algorithms \cite{girdhar_autonomous_2014} have the challenge of sampling in real-time without advance knowledge of the upcoming visual data or the length of the expedition. Instead, they can only observe one frame at a time and compute the features solely for that specific frame. Therefore, sampling in ON-IS algorithms involves comparing between the information from previously seen frames and the current frame. In this paper, we focus on ON-IS algorithms as they are robust to various real-time operations and can potentially be used in an offline manner as well.

Our proposed approach, Semantic Online Informative Sampling (SON-IS), is inspired by the Real-time Online Spatiotemporal Topic modeling (ROST) \cite{girdhar_autonomous_2014, girdhar_efficient_2012, Girdhar2014} algorithm which is an online informative sampling approach for improving the exploration capability of AUVs. This algorithm samples visual frames which are significantly more distinct from the previous frames observed until the current timestep. This helps an AUV navigate toward unseen areas for exploration and to discover new observations. However, this may not be as useful for monitoring tasks since visual uniqueness does not necessarily imply semantic uniqueness of a frame. For example, scenes with the same fishes from different perspectives or in different scenarios may be visually unique but they are not unique semantically. Thus, we propose to encode semantic features into the informative sampler with the intention of identifying semantically unique frames. As such, SON-IS can significantly reduce the search time of interesting objects in long visual feeds. For example, SON-IS can help a marine biologist quickly identify all the unique marine life captured from a long sequence of underwater monitoring visual data.

Based on the success of many OFF-IS methods \cite{otani2016video, wang2014real, jiang_comprehensive_2019}, we propose to use features from the semantic object detection as the input features for the online informative sampler described in ROST \cite{girdhar_autonomous_2014, girdhar_efficient_2012}. 
This encodes the semantic information within the features which in turn improves the robustness of informative sampling. Our contributions are as follows: 
\begin{itemize}
    \item a novel approach (SON-IS) for semantically-aware online informative sampling which combines an object detection model with an online semantic sampler
    \item a novel evaluation metric to validate the semantics of the automated samples against the human-picked samples
    \item a robust validation of the proposed algorithm using human user studies and real-world data
\end{itemize}


\section{Related Work}

\subsection{Informative Sampling}
The informative sampling problem has been explored for both online and offline sample generation in  \cite{girdhar_autonomous_2014, otani2016video, wang2014real, jiang_comprehensive_2019}. We discuss the details of these papers in the following section.

\subsubsection{Online Sampling}
The state-of-the-art work on online sampling approach is referred as ROST and proposed by Girdhar et al. \cite{girdhar_autonomous_2014, girdhar_efficient_2012, Girdhar2014}.
ROST utilizes semantic-agnostic features such as ORB \cite{rublee2011orb}, Color, Gabor, and Texton (texture-based) features. These features are transformed into bag-of-words representation using a vocabulary and fed into a topic model that groups related features in spatio-temporal neighborhoods under the same topics in an unsupervised manner. By comparing the topic distribution between the current chosen samples and the next one in real-time, ROST determines the surprise factor, which is used to update the online samples. Another ON-IS approach is \cite{lal_online_2019}, which proposes a convolutional Long Short-Term Memory (LSTM) network to encode both spatial and temporal features. However, it is a supervised method which was designed and evaluated for online video summarization which requires the information about the video length and subtitles in advance for training and evaluation. Such information is not readily available in our proposed application of underwater environment monitoring. In our proposed approach, we build upon the ROST framework which allows us to perform real-time sampling in underwater environments. 

\subsubsection{Offline Sampling}
In contrast to ON-IS, \cite{otani2016video} and \cite{10.1007/978-3-319-10584-0_33} leverage semantic features for dynamic informative sampling, but their approach is not specifically designed for online purposes. They adopt a clustering-based sampling technique where learning based features are extracted from each original video segment to generate the video summary. Despite their effectiveness, these methods lack the real-time adaptability. Jiang et al. \cite{jiang_comprehensive_2019} propose a scalable deep neural network for informative sampling in a content-based recommender systems formulation. Their approach predicts the segment's importance score by considering both segment and visual data level information. However, this approach is not optimized for real-time applications. Additionally, in \cite{sciencedirectVSUMMMechanism}, VSUMM is presented as a methodology for generating static video summaries by utilizing color feature extraction and the k-means clustering algorithm. The authors also propose an approach for evaluating video static summaries, where user-created summaries are compared to the proposed approach. Similarly in \cite{6786125}, they also used color features to generate static summaries by proposing a modified hierarchical clustering approach. Furthermore, \cite{MEI2015522} addresses the informative sampling task with a minimum sparse reconstruction (MSR) problem. The objective is to reconstruct the original video sequence with as few selected key frames as possible. The authors propose an efficient and effective MSR-based informative sampling method suitable for both offline and online applications. Another OFF-IS algorithm is \cite{wang2014real} which introduces informative sampling for user-generated videos based on semantic recognition. In terms of speed efficiency, it is able to produce a summary for a $2$ minute video in $10$ seconds, but it takes the entire video as input before producing a summary. Thus, there is a high requirement for memory and the duration of summarization would only increase as the input video gets longer. 

Overall, while existing approaches demonstrate promising results for informative sampling, our proposed framework aims to address real-time adaptability while ensuring semantic content in the selected samples, which has not been fully explored in the current literature. In addition, we propose an evaluation metric to assess the outcome of informative sampling approaches.

\subsection{Semantic Recognition} 
An offline semantic recognition algorithm is proposed in \cite{wang2014real} to identify semantic events occurring in video sequences. The semantic recognition module is based on the work in \cite{jiang2012super}. Our proposed approach, in contrast, detects scene semantics in real-time for each frame using an object detector module and selects only the most unique visual frames using an online informative sampler.

We chose DETR\cite{carion2020endtoend} for semantic feature extractor as it is based on the Transformer architecture and offers a number of advantages over the state-of-the-art Faster-RCNN \cite{ren2016faster} algorithm in terms of the throughput and faster training time. The Transformer-based framework \cite{vaswani2017attention} of DETR introduces a novel approach that directly predicts object class labels and bounding box coordinates for all objects present in an image. The model's encoder is composed of a Convolutional Neural Network (CNN) backbone such as ResNet \cite{he2015deep} for feature extraction, combined with a stack of transformer layers to capture long-range dependencies and contextual relationships between objects. DETR's transformer layers leverage self-attention mechanisms to attend to different regions in the input, allowing the model to consider global context while making predictions. During training, DETR employs a bipartite matching loss that integrates class prediction and box regression losses, enabling accurate and robust object detection. Leveraging the powerful feature extraction capabilities of DETR, our proposed framework incorporates the semantic feature extraction module which is fine-tuned on an underwater dataset to learn context-aware visual representations for effective underwater semantic informative sampling.

\section{Proposed Methodology}

\begin{figure*}[htb]
  \centering
    \includegraphics[width=\textwidth]{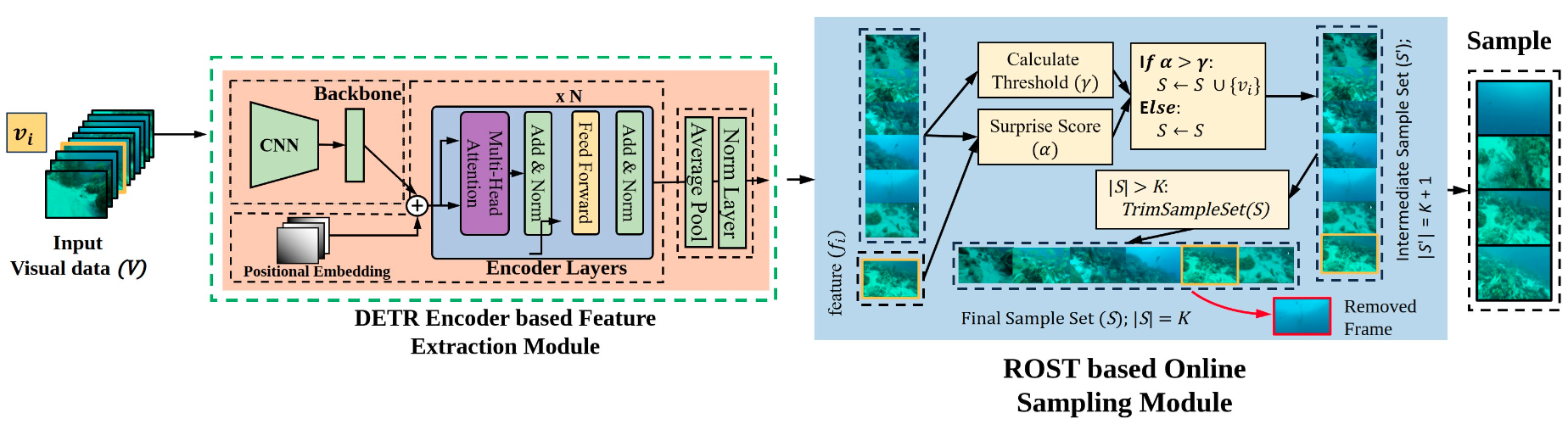} 
    \caption{Overview of SON-IS algorithm comprising of the feature extraction and online sampling modules.}
    \label{fig:rost_and_proposed_framework}
\end{figure*}

The overview of our proposed Semantic Online Informative Sampling (SON-IS) approach is illustrated in Fig. \ref{fig:rost_and_proposed_framework}. Specifically, we use DETR as the feature extraction module and based our online sampler module on ROST's sampler. The details of the two modules are presented in this section.

\subsection{Feature Extraction Module}
\label{Method: Semantic Features}
Our feature extraction module comprises an encoder which is obtained from DETR \cite{carion2020endtoend}. DETR consists of a CNN backbone and a transformer \cite{vaswani2017attention}. The CNN backbone is a ResNet \cite{he2015deep} model pre-trained on the ImageNet \cite{deng2009imagenet} dataset for classification tasks. Together with the transformer encoder-decoder, it is subsequently trained on the COCO \cite{lin2015microsoft} dataset, as described in \cite{carion2020endtoend}, to acquire rich semantic features crucial for object detection. Utilizing this pre-trained model, we fine-tuned the encoder and decoder components of the pre-trained model by training the model on an underwater dataset while keeping the backbone part frozen. This helped the model to learn features specific to underwater objects, allowing it to better learn the semantics of an underwater environment. Subsequently, we utilize only the CNN backbone and the encoder to extract features from given visual data for semantic sampling. 
The features from the encoder capture various levels of visual information and contextual details and are used as semantic information.

In our proposed framework, at time $t$, the input visual frame $v_t \in \mathbb{R}^{C \times H \times W}$ is first passed through the CNN backbone $\mathbf{F}_b$. Position embeddings $\mathbf{p}_e$ are added to these features, which are then passed through the DETR encoder $\mathbf{F}_e$. Consequently, the DETR-based approach produces features $f_t$ at time $t$, computed as $f_t = \mathbf{F}_e (\mathbf{F}_b (v_t) + \mathbf{p}_e)$. 
Once we obtain the feature $f_t$, we then use average pooling to reduce the dimensions of $f_t$ while capturing the essence of each feature. Specifically, we reduce from $256$$\times$$m$$\times$$n$ where $m$$\times$$n$ is proportional to the dimension of the visual data to $256$$\times$$1$$\times$$1$. This reduction in spatial dimensionality merges all the cases where the same fish appears in different locations of an image to be of a similar feature. Then, the features are normalized by the maximum value of $f_{t}$ so that the values are between $0$ and $1$. We then pass the features to our online informative sampling module. 

\subsection{Online Informative Sampling Module}
\label{sec:onvs_formulation}

Online informative sampling algorithms aim to create samples $\mathcal{S}$ from a given long visual sequence $V$ of length $N$ (not known in advance) by selecting the most informative frames and removing redundant ones. The set of visual data is denoted as $V = \{v_i, i \in N\}$ where $v_i \in \mathbb{R}^{C \times H \times W}$, $C$ is number of color channels, $H$ is height of image and $W$ is its width. The ON-IS algorithms consist of two main components: a feature extractor $\mathbf{F}$ and a sampler $\mathbf{S}$ as shown in Fig. \ref{fig:rost_and_proposed_framework}. These algorithms also adopt a pre-determined sample size denoted as $K$, where $K << N$, to suit its use case. 

The state-of-the-art ON-IS algorithm, ROST \cite{girdhar_autonomous_2014}, computes the features $f_i$ for the first $K$ frames individually, $f_i = \mathbf{F}(v_i); i \in \{1, 2, \cdots, K\}$, and puts them into the sample set (summary) $\mathcal{S}$, serving as the prior sample set for subsequent frames. At frame $t > K$, given the current sample set $\mathcal{S} = \{{s_i}\}_{i=1}^{K}$, the algorithm determines the surprise score $\alpha$ of the new frame ${v_t}$,
 which is the shortest distance (Symmetric KL-divergence) from the current frame to the sample set. Additionally, to decide whether to include the current frame in the sample set, a threshold $\gamma$ is used. This threshold is set as the mean of the shortest distance between each sample frame and all other sample frames. 
Now, for each new frame obtained at timestamp $t$, its score is computed given the current sample set, and if the score is above the threshold $\gamma$, it is added to the current sample set. This algorithm is called Picking Above The Mean \cite{Girdhar2014}. This maximizes the distance between the samples chosen.
To keep the sample set within sample size $K$, the least informative sample is removed from the set $S$, by re-performing the greedy k-centers clustering algorithm \cite{Girdhar2014}. This trimming process ensures that the computational cost of the algorithm is low. 

In the context of our proposed SON-IS algorithm, we obtain semantic features that are distinct for different classes of objects to input them into the online informative sampling module. This allows the online sampling module to select frames that are not only representative of the visual content but also semantically attuned with the desired target objects.

\subsection{Evaluation Metric}
We propose a novel evaluation metric named "Semantic Representative Uniqueness Metric (SRUM)" that scores the semantic meaningfulness as well as representativeness of the automated samples by comparing them with human-picked samples. To do so, we manually label both the human and automated samples with the name of the desired target object (e.g., fish species) identified in the frame. 
If a frame consists of more than a single species of fish, we label the frame with a list of all observed fishes. Once the frames are labelled, a matching process described in Algorithm \ref{alg:semantic_score_alg} is conducted to match the automated sample frames to human-picked samples. This metric can be applied to any informative sampling approach.

\begin{algorithm}
\caption{  Semantic Representative Uniqueness Metric (SRUM) ($A$, $H$, $L$, $fps$, $\alpha$)}\label{alg:semantic_score_alg}
\begin{algorithmic}
\State $overall\_score \gets 0$
\For{$i \gets 0$ to $length(H)$}
    \State $Score\_list \gets [ ]$
    \For{$j \gets 0$ to $length(A)$}
        \State $Human\_label \gets L[H_i]$
        \State $Automated\_label \gets L[A_j]$
        \For{$k \gets 0$ to $length(Automated\_label)$}
            \If{$Automated\_label_k$ in $Human\_label$}
                \State $Representative\_score \gets RS(H_i, A_j, fps)$
                \State $Semantic\_score \gets 1$
                \State $score = $\parbox[t]{.6\linewidth}{
                $\alpha Semantic\_score \\
                + (1 - \alpha)Representative\_score$
                }
                \State $Score\_list.insert(score)$
            \EndIf
        \EndFor
    \EndFor
    \State $max\_score \gets max(Score\_list)$
    \State $overall\_score = overall\_score + max\_score$
\EndFor
\State \Return {$\frac{overall\_score}{length(H)}$}

\end{algorithmic}
\end{algorithm}

The SRUM metric accounts for both the Semantic Score and the Representative Score, which is explained below. In Algorithm \ref{alg:semantic_score_alg}), $A$ represents the frames of a sample selected by an automated sampler, $H$ represents the frames selected by a particular human subject, $L$ represents the dictionary which correlates frames with their semantic labels,  $fps$ represents frames per second of the visual feed, and $\alpha$ represents the weight of the Semantic Score, where $(1 - \alpha)$ is conversely the weight of the Representative Score. 

The Semantic Score rewards a particular sample frame for having a semantic label that can also be found in the list of labels of any frames of the human-picked sample. When this happens, it is a match and a Semantic Score of $1$ is given to the particular automated frame. Otherwise, the Semantic Score is $0$.

To mitigate the problem of duplicated matches, the SRUM metric gives a score to each human-picked frame rather than to each automated sample frame (scoring from perspective of human sample). This scoring allows the matching of many human frames to one automated frame, but not many automated frames to one human frame, to mitigate the case where an automated sample contains $6$ exact same frames which match to one human frame and getting a perfect score for that. Conversely, we assume the human-picked samples to be unique and representative, and even in the case of duplicated human sample frames, it is favorable and further denotes the representativeness or uniqueness of the duplicated frames.

To further evaluate how representative the chosen automated sample frame is for that particular set of fish species, we further score it based on the distance in time the automated sample is from the matched human sample. We call this the Representative Score, which is computed using Equation \ref{alg:representative_score} where $fps$ represents the frames per second of the visual feed. 
\begin{equation}
\label{alg:representative_score}
RS(H_i, A_j, fps) =  
\begin{cases}
    1,      & \text{if } e^{-\frac{|A_j - H_i|}{fps} + 2} \geq 1\\
    e^{-\frac{|A_j - H_i|}{fps} + 2},              & \text{otherwise}
\end{cases}
\end{equation}
A perfect score of $1$ is given if the distance between the automated and human frame is less than $2$ seconds and exponentially decreases for distances greater than $2$ seconds. This threshold was chosen following the threshold value used in \cite{MEI2015522}. It should also be noted that the Representative Score is only given if there is a semantic match between the labels, i.e. even if two frames are close in frame number distance, the score would be $0$ if there are no matches in the labels. This is done to penalize the lack of a relevant or matching fish in a frame, e.g. if in frame $10$ there is no fish, but in frame $11$ there is a "Clown-fish", then frames $10$ and $11$ do not match at all since the "Clown-fish" species is not sufficiently visible in frame $10$.

\section{Experimental Setup}
\subsection{Finetuning Dataset}
\label{sec:fintuning_dataset}
We fine-tuned the encoder and decoder components of the pre-trained DETR model using the Brackish Underwater Object Detection dataset \cite{pedersen2019brackish}. This dataset comprises 14,674 images, of which 12,444 images are annotated with bounding boxes, depicting various marine creatures such as fish, crabs, and other marine animals. The data was captured using a static camera installed at a depth of 9m beneath the Limfjords bridge situated in northern Denmark. This dataset is already split into three parts: training, validation, and testing, containing 11,739, 1,467, and 1,468 images, respectively. All the images used in the experiments are of size 1920x1080.

\subsection{Evaluation dataset}
\begin{figure}[h]
    \centering
    \includegraphics[width=\linewidth]{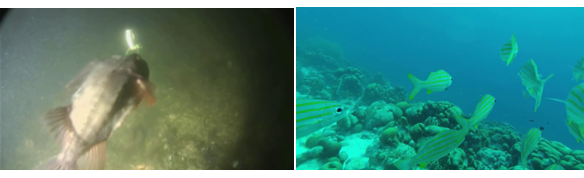} 
    \caption{The left image is from the Brackish dataset \cite{pedersen2019brackish} used for fine-tuning and right image is from the evaluation dataset \cite{solodive}.} 
    \label{fig:train_eval_dataset}
\end{figure}

To evaluate our proposed SON-IS algorithm, we used a publicly available underwater diving video \cite{solodive} sourced from YouTube. We input only one frame at a time from start to end without \textit{a priori} knowledge of the length of the video. 
We chose this video because it is captured with a moving camera which is similar to the visual data collection process by underwater vehicles or humans for persistent monitoring or exploration tasks. However, the fine-tuning training data was collected using a static camera. The evaluation data is also visually significantly different from the fine-tuning dataset. Fig. \ref{fig:train_eval_dataset} shows a comparison of the sample images from each dataset.

Specifically, the fine-tuning training dataset comprises a small variety of fishes whereas, the evaluation dataset comprises a wide array of fish species. 
The evaluation dataset also has more variety in color and shapes compared to the Brackish fine-tuning dataset. Thus, this out-of-distribution dataset serves to test the robustness of our proposed (SON-IS) approach. 

\section{Results and Discussion}
\subsection{Collection of Human-picked Samples}
To evaluate the automated samples, human-picked samples from $6$ subjects were gathered as the ground-truth for the sampling dataset. Since the objective was to capture as many unique species of aquatic animals as possible in $6$ frames, the human subjects were informed beforehand to summarize ``interesting events related to fish". To mimic an online informative sampling process, the human subjects were asked to make a $6$-frame sample of the video without knowing its duration. However, they were allowed to select and save multiple potential sample frames while watching the video. Afterward, they chose the final $6$ images from the saved potential images without rewinding the video. 

\subsection{Qualitative Evaluation}
\begin{figure*}[t!]
  \centering
    \includegraphics[width=\textwidth]{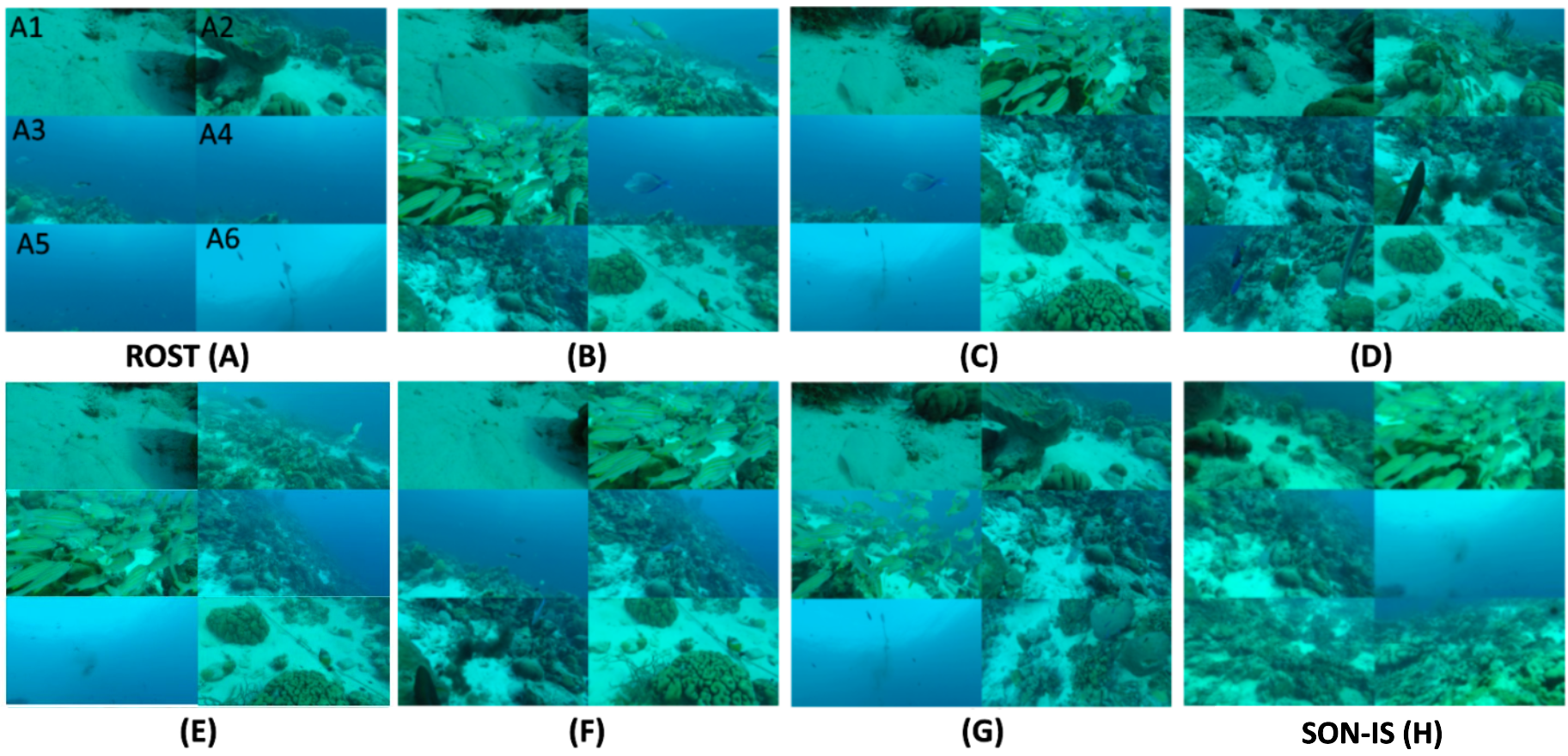} 
    \caption{Qualitative study of ROST sample ($A$), SON-IS sample ($H$, Our Method) and human-picked samples ($B$, $C$, $D$, $E$, $F$ and $G$) .}
    \label{fig:summary_comparison}
\end{figure*}

The output samples from state-of-the-art (ROST) and our proposed algorithm (SON-IS), along with six human-picked samples, are presented in Fig. \ref{fig:summary_comparison}. Upon comparing the samples, we observed that participants $B$, $C$, $D$ and $F$ selected many similar frames, with at least $5$ out of the $6$ samples having the same fish. On the other hand, participants $E$ and $G$ selected $4$ frames with similar fishes, and $2$ frames each with no matching fish. Qualitatively, we found that our proposed method captured more frames with unique species of fish compared to ROST.

Table \ref{tab:cifar100_class_incre} provides information on the frames selected by different methods corresponding to Fig. \ref{fig:summary_comparison}. Each row represents a different method of sampling, and each column ($1$ to $6$) corresponds to the frame number of the selected sample in respective order in the sample set. 
\vspace{-0.01\textheight}
\begin{table}[!htb]
\caption{Selected sample frames by all methods. Human sample frame numbers having SRUM matches with SON-IS are highlighted}
    \begin{center}
        \label{tab:cifar100_class_incre}
        {\footnotesize
        \tabcolsep=0.17cm
        \begin{tabular}{{l}{r}*6{r}} 
        \hline
        \textbf{Methods} & \textbf{1} & \textbf{2} & \textbf{3} & \textbf{4} & \textbf{5} & \textbf{6} \\
        \hline
        ROST (A) & 182 & 600 & 4326 & 4484 & 4566 & 6736 \\
        Human subject (B) & \textbf{242} & 1897 & \textbf{3003} & 4454 & \textbf{6421} & 12104 \\
        Human subject (C) & \textbf{294} & \textbf{2980} & 4438 & \textbf{6461} & \textbf{6711} & 12064 \\
        Human subject (D) & \textbf{444} & \textbf{2820} & \textbf{6465} & 6507 & \textbf{9536} & 12083 \\
        Human subject (E) & \textbf{191} & 1803 & \textbf{3015} & \textbf{6207} & \textbf{6625} & 12082 \\
        Human subject (F) & \textbf{136 }& \textbf{3022} & 4281 & \textbf{6357} & 6500 & 12061 \\
        Human subject (G) & \textbf{293} & 597 & \textbf{3268} & \textbf{6466} & \textbf{6716} & \textbf{8082} \\
 
        \hline
        SON-IS (H) & \textbf{538} & \textbf{2996} & \textbf{6404} & \textbf{6630} & \textbf{7578} & 9104 \\
        \hline
    \end{tabular}}
    \end{center}
\end{table}

\vspace{-0.02\textheight}
\subsection{Quantitative Evaluation}
We use SRUM metric for the quantitative evaluation, where the final score for the automated sample is its average SRUM score across all human-picked samples. Table \ref{tab:quantitative_eval} shows the quantitative scores for sampling methods $A$ and $H$ compared with the human samples $B$, $C$, $D$, $E$, $F$ and $G$ using the SRUM metric. Using varying values of $\alpha$ ($\alpha = 0.5, 0.75, 1.0$), SON-IS outperforms ROST for these $\alpha$ values, showing that our proposed method is able to better identify semantically meaningful sample frames. These $\alpha$ values are chosen so that the weight for Semantic Score is equal to or higher than Representative Score, in order to evaluate the semantic meaningfulness of the automated samples.
\vspace{-0.005\textheight}
\begin{table}[H]
\caption{Quantitative Comparison of the ROST sample (A) and SON-IS sample (H) using the average SRUM score against all human-picked samples.}
    \begin{center}
        \label{tab:quantitative_eval}
        {\footnotesize
        \tabcolsep=0.17cm
        \begin{tabular}{l|{r}{r}|r} 
        \hline
        \textbf{Alpha} & \textbf{ROST} & \textbf{SON-IS} & \textbf{Human-picked Samples} \\
        \hline
        0.5 & 0.261 & \textbf{0.460} & 0.525 \\
        0.75 & 0.297 & \textbf{0.535} & 0.601 \\
        1.0 & 0.333 & \textbf{0.611} & 0.677 \\
        \hline
    \end{tabular}}
    \end{center}
\end{table}
We also evaluated the human-picked samples against one another using the SRUM metric, as if each human sample being evaluated is treated as an automated sample (and removed from the pool of human samples). We computed the mean of the average SRUM score of all human samples against the rest and recorded it under the "Human-picked Samples" column. This gives an idea of how a human sample scores with respect to other human samples so as to provide a benchmark for the score of the automated samples. Then, we took the "Human-picked Samples" benchmark and recalculated ROST and SON-IS scores as a percentage of the mean human score, and recorded it in Table \ref{tab:percentage score}, which shows their performance judged with respect to human samples. While ROST's sample is, on average, $49.4\%$ as good as human samples, SON-IS is $89\%$ as good as human samples. 
\vspace{-0.005\textheight}
\begin{table}[H]
\caption{Percentage score relative to mean human score}
    \begin{center}
        \label{tab:percentage score}
        {\footnotesize
        \tabcolsep=0.7cm
        \begin{tabular}{{l}{r}*6{r}} 
        \hline
        \textbf{Alpha} & \textbf{ROST} & \textbf{SON-IS} \\
        \hline
        0.5 & 49.7\% & \textbf{87.6\%}  \\
        0.75 & 49.4\% & \textbf{89.1\%} \\
        1.0 & 49.2\% & \textbf{90.2\%} \\
        Average & 49.4\% & \textbf{89.0\%} \\
        \hline   
    \end{tabular}}
    \end{center}
\end{table}

\vspace{-0.01\textheight}
\section{Conclusion}
Our research has yielded compelling results through a cross-comparison between a preliminary human study and experiments with the state-of-the-art informative sampling approach, ROST, and our proposed Semantic Online Informative Sampling (SON-IS) approach. Qualitatively, we observe that our approach excels in capturing semantically meaningful samples of underwater visual data when compared to ROST. Quantitatively, SON-IS outperformed ROST using the SRUM metric for different $\alpha$ values. For future work, validation could be performed on more visual sequences and on other evaluation metrics.

\section*{Acknowledgement}
This project is supported by A*STAR under its RIE2020 Advanced Manufacturing and Engineering (AME) Industry Alignment Fund (Grant No. A20H8a0241).

\bibliographystyle{IEEEbib}
\bibliography{ocean_final_2024}
\end{document}